\documentclass[letterpaper, 10 pt, conference]{ieeeconf}  % Comment this line out if you need a4paper

\IEEEoverridecommandlockouts                              % This command is only needed if
\usepackage{hyperref}
\usepackage{threeparttable}
\usepackage{graphicx}
\usepackage{multirow}
\usepackage{times}
\usepackage{latexsym}
\usepackage{xurl}
\usepackage{xcolor}
\usepackage{color} %to define color
\definecolor{darkgreen}{rgb}{0.44,0.68,0.28}
\definecolor{darkblue}{rgb}{0,0.44,0.75}
\definecolor{mediumgrey}{rgb}{0.65,0.65,0.65}

\title{\LARGE \bf
Rare Disease Identification from Clinical Notes with Ontologies and Weak Supervision
}

\author{Hang Dong, Víctor Suárez-Paniagua, Huayu Zhang, Minhong Wang, Emma Whitfield, and Honghan Wu% <-this % stops a space
\thanks{This work is supported by Health Data Research UK National Phenomics and Text Analytics Implementation Projects and Wellcome Institutional Translation Partnership Awards (PIII009, PIII029, PIII032, PIII054).}
\thanks{Hang Dong, Víctor Suárez-Paniagua, Minhong Wang are with Centre for Medical Informatics, Usher Institute, University of Edinburgh, Edinburgh, UK. Huayu Zhang is with Advanced Care Research Centre, Usher Institute, University of Edinburgh, Edinburgh, UK. Emma Whitfield and Honghan Wu are with Institute of Health Informatics, University College London, London, UK. Hang Dong, Víctor Suárez-Paniagua, Emma Whitfield and Honghan Wu are also with Health Data Research UK, London, UK.}
\thanks{Corresponding author: Hang Dong ({\tt hang.dong@ed.ac.uk})}
}

\begin{document}

\maketitle
\thispagestyle{empty}
\pagestyle{empty}

%%%%%%%%%%%%%%%%%%%%%%%%%%%%%%%%%%%%%%%%%%%%%%%%%%%%%%%%%%%%%%%%%%%%
\begin{abstract}

The identification of rare diseases from clinical notes with Natural Language Processing (NLP) is challenging due to the few cases available for machine learning and the need of data annotation from clinical experts. We propose a method using ontologies and weak supervision. The approach includes two steps: (i) Text-to-UMLS, linking text mentions to concepts in Unified Medical Language System (UMLS), with a named entity linking tool (e.g. SemEHR) and weak supervision based on customised rules and Bidirectional Encoder Representations from Transformers (BERT) based contextual representations, and (ii) UMLS-to-ORDO, matching UMLS concepts to rare diseases in Orphanet Rare Disease Ontology (ORDO). Using MIMIC-III US intensive care discharge summaries as a case study, we show that the Text-to-UMLS process can be greatly improved with weak supervision, without any annotated data from domain experts. Our analysis shows that the overall pipeline processing discharge summaries can surface rare disease cases, which are mostly uncaptured in manual ICD codes of the hospital admissions.
\newline

\indent \textit{Clinical relevance}— The text- and ontology-based approach can largely reduce missing cases in rare disease cohort selection.
\end{abstract}

%%%%%%%%%%%%%%%%%%%%%%%%%%%%%%%%%%%%%%%%%%%%%%%%%%%%%%%%%%%%%%%%%%%%%%%%%%%%%%%%
\section{Introduction}
\label{intro}
Rare diseases are those that affect 5 or fewer people in 10,000; there are between 6,000 and 8,000 rare diseases and they are collectively essential, e.g. affecting approximately 8\% of the population in Scotland \cite{scot_gov_2021}. Compared to common diseases, rare diseases are less likely being explicitly coded because they are under-represented in the current, ICD-10 (International Classification of Diseases, version 10) terminologies \cite{Bearryman2016}. In terms of automated coding, most existing Natural Language Processing (NLP) models (especially with deep learning) perform well on the most common diseases but tend to fail on the long tail of infrequent diseases due to the lack of instances for training \cite{rios-kavuluru-2018-shot,dong2021}. Also, these NLP models usually rely on the potentially incomplete codes as gold standard and overlook the large set of under-coded cases \cite{searle2020}, e.g. for rare diseases.

The key challenge for rare disease identification with NLP is the lack of annotated data. Annotating a variety of rare diseases in clinical notes from scratch requires specific domain expertise and a very large corpus (to have enough cases), thus considerable cost and time from a group of clinical experts. This study proposes a practical and effective approach leveraging ontologies and weak supervision to alleviate the burden of annotation.

\begin{figure}
  \centering
  \includegraphics[width=0.48\textwidth]{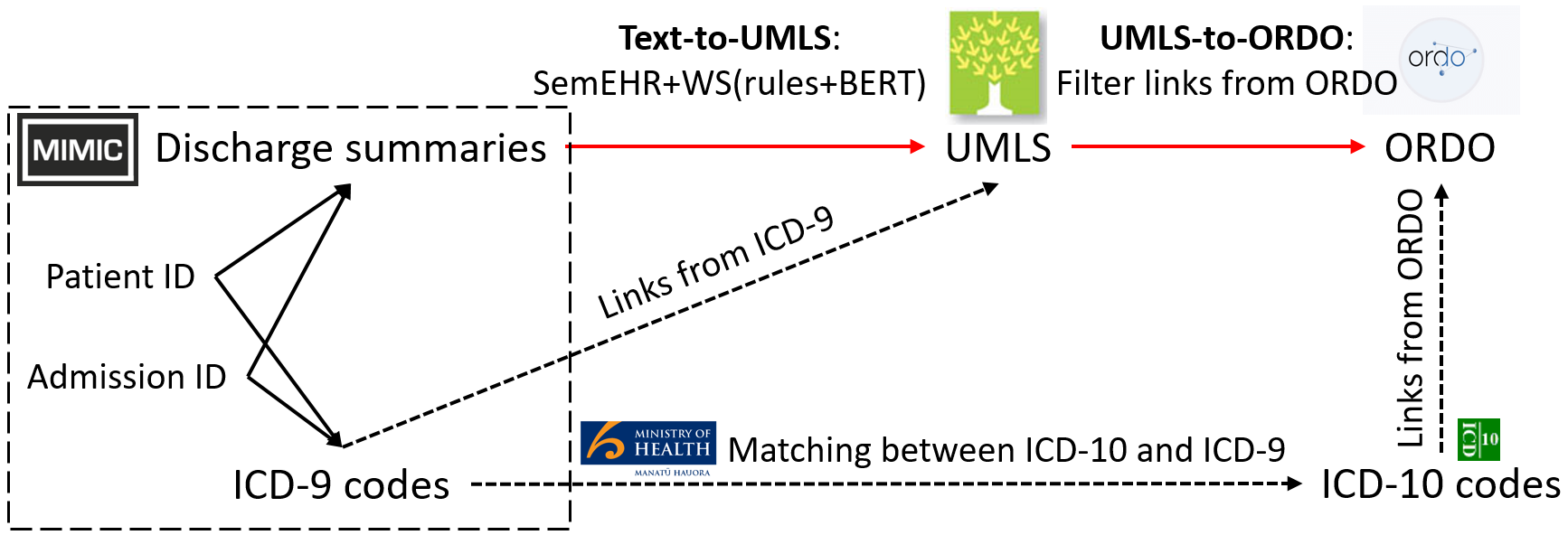}
  \caption{A pipeline for rare disease identification from clinical notes with ontologies and weak supervision. The upper horizontal lines (in \textcolor{red}{red}) show the proposed pipeline based on clinical notes (e.g. discharge summaries) and ontologies, including two steps (Text-to-UMLS and UMLS-to-ORDO). No annotation data are needed, through a UMLS extraction tool, SemEHR, and weak supervision (WS) based on customised rules and BERT-based contextual representations (see details on WS in Fig. \ref{pipeline-weak}). The lower, dotted lines show a baseline approach purely based on manual ICD codes, also enhanced with ontology matching.}\label{pipeline-main}
\end{figure}

Ontologies provide viable sources of knowledge as concepts and relations to estimate rare diseases from clinical notes \cite{kahn2017ontology}. In this work, we cast rare disease identification as entity linking and ontology matching problems, i.e. linking positive text mentions from clinical notes to concepts in clinical ontologies. We match mentions to Orphanet Rare Disease Ontology (ORDO) \cite{weinreich2008orphanet} and use Unified Medical Language System (UMLS) as an intermediary dictionary to extend matching terms. The approach is thus comprised of two integrated parts, as shown in Fig. \ref{pipeline-main}: (i) Text-to-UMLS, i.e. UMLS concept identification from texts through weak supervision and (ii) UMLS-to-ORDO, i.e. ORDO concept identification through ontology matching.

\begin{figure*}[ht]
  \centering
  \includegraphics[width=0.8\textwidth]{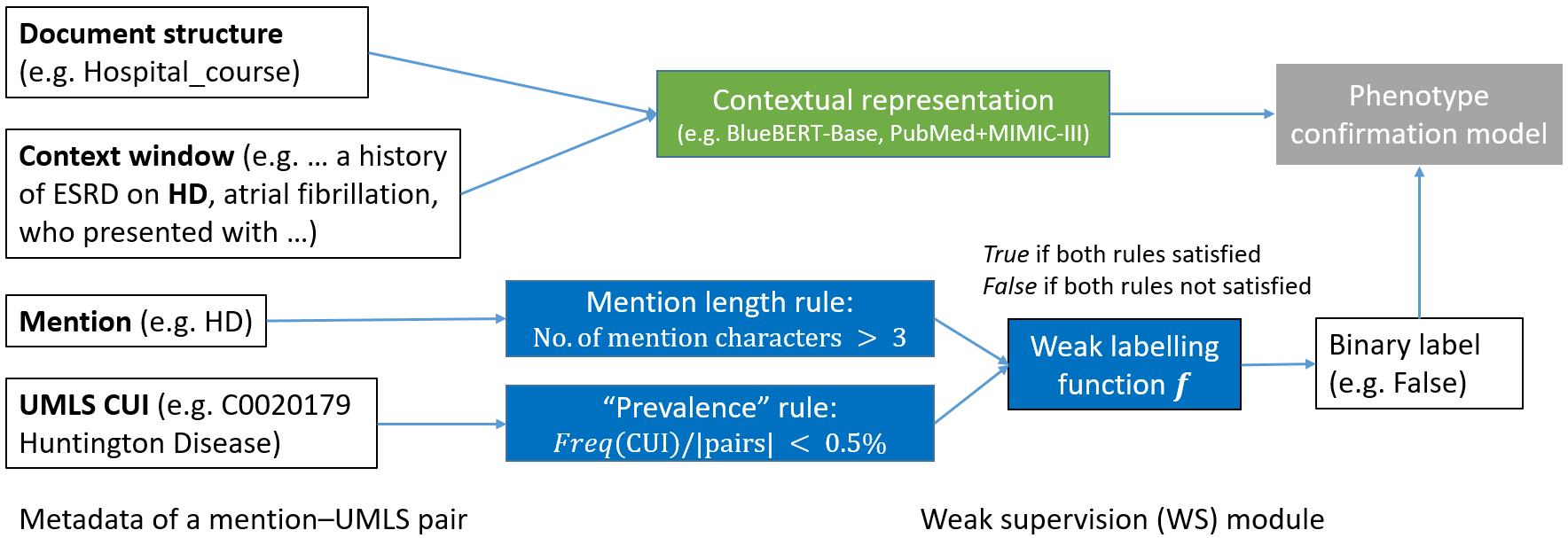}
  \caption{Weak supervision process for Text-to-UMLS linking. The left four white text boxes displayed the metadata (with examples) of a candidate mention-UMLS pair, identified by SemEHR \cite{Wu2018semehr}; the coloured text boxes in the middle show the contextual representation block (in \colorbox{darkgreen}{\textcolor{white}{green}}) and the rule-based weak data labelling process (in \colorbox{darkblue}{\textcolor{white}{blue}}). A binary label is then generated, which weakly estimates whether the candidate pair indicates a correct phenotype of the patient. A phenotype confirmation model (in \colorbox{mediumgrey}{\textcolor{white}{grey}}) can then be learned to select correct phenotypes from the pairs.}\label{pipeline-weak}
\end{figure*}

Weak supervision \cite{wang_clinical_2019,ratner2019} is a strategy to automatically create weakly labelled training data using heuristics, knowledge bases, crowdsourcing, and other sources, to alleviate the burden and cost of annotation. In clinical NLP, previous studies proposed using lexical or concept filtering rules to create weakly labelled data, then usually represented with neural word embeddings (and more recently with BERT \cite{devlin-etal-2019-bert}), to classify clinical texts \cite{wang_clinical_2019} into nuanced categories, e.g. suicidal ideation \cite{cusick2021} or lifestyle factors for Alzheimer's Disease \cite{shen2021}. We extend the weak supervision process with customised rules to refine a named entity linking tool to identify rare diseases. For Text-to-UMLS linking (see Fig. \ref{pipeline-main}-\ref{pipeline-weak}), we propose to efficiently create weak training data of sufficient quality (candidate mention-UMLS pairs) with two rules, mention character length (regarding ambiguous abbreviations) and ``prevalence'' (regarding rare diseases), on top of a gazetteer-based named entity linking tool, SemEHR \cite{Wu2018semehr}. We further applied a clinically pre-trained BERT \cite{peng2019transfer} to capture the contexts under-lied in the texts to disambiguate the mention to improve entity linking.

% an illustration of the data creation process

%to trim
We tested the whole pipeline using discharge summaries (n=59,652) in the MIMIC-III dataset \cite{johnson_mimic-iii_2016} and created a large entity linking dataset (of 127,150 candidate mention-UMLS pairs) for weak supervision. The weakly trained model was then evaluated on a gold-standard annotated data of 1,073 mentions. The weak supervision with contextual embedding dramatically improved the precision and $F_1$ scores of SemEHR with minimum loss of recall, and significantly improved the approach using the rules alone with the tool. The proposed approach adapted for rare diseases also outperformed the recent off-the-shelf tool, Google Healthcare Natural Language API \cite{Bodnari2020}. Leveraging the linkage from ICD-9 codes to ORDO concepts, we further show that the rare disease admission cases identified from discharge summaries can greatly complement the results from the manual ICD codes. The overall pipeline has the potential to prevent missing cases in rare disease cohort selection.

\section{Method}
%would need a better figure to demonstrate the weak supervision process - could be used in a journal paper later - done
% illustrating weak supervision, seen, unseen, ...
%We thus describe the main design of the system to identify rare diseases from texts.
%We aim to design a that can identify rare disease mentions from clinical notes.
%The main idea to identify rare disease from clinical notes is to use ontology-based entity linking, linking free-text mentions to concepts in ORDO, as shown in Fig. \ref{pipeline-main}. Weak supervision with contextual representation is proposed to improve the matching from free-text mentions to disease concepts in UMLS.
We describe both the ontology-based approach and the weak supervision process below.

\subsection{Ontology-based Extraction and Linking}
Rare diseases cover a wide range of human phenotypes, curated in clinical ontologies. ORDO \cite{weinreich2008orphanet} contains the most comprehensive set of rare diseases with external links to UMLS, ICD-10, and other ontologies. ORDO was used in \cite{kahn2017ontology} to estimate the frequency of rare diseases in radiology reports. We further extend this idea by using UMLS as an intermediary dictionary to leverage its rich synonyms to improve the coverage of matching to texts.

We applied a UMLS concept extraction tool, SemEHR, which has been deployed in health data safe havens and private servers in the UK \cite{Wu2018semehr}. SemEHR adapts Bio-YODIE \cite{gorrell2018} as its main NLP module with further learning functionalities, e.g. rule- and ontology-based learning, and a search interface. Bio-YODIE can efficiently extract UMLSs from texts using a gazetteer-based matching approach. One issue of both tools, however, is that they assume a strong prior to assign the same UMLS to the mention regardless of its context \cite{gorrell2018}. This can results in many false positive phenotypes, especially for abbreviations used in clinical notes. For example in Table \ref{fp_example}, none of the identified ``HD'' mentions contained in the sections indicate a type of disease.

\begin{table*}[th]
\caption{Examples of false positive phenotypes in the candidate mention-UMLS pairs from SemEHR and Bio-Yodie}
%\small
\center
\label{fp_example}
\begin{tabular}{ll||ll}
\cline{1-4}
\textbf{Mention} in context window                            & Meaning                    & UMLS by SemEHR                                                                                 & UMLS by Bio-YODIE \\
\cline{1-4}
%\textbf{Ambiguous abbreviation, e.g. HD} & & \\
His temporary \textbf{HD} line was pulled.                    & Medical device              & \multirow{4}{*}{\begin{tabular}[c]{@{}l@{}}Hundington Disease\\(CUI: C0020179)\end{tabular}} & \multirow{4}{*}{\begin{tabular}[c]{@{}l@{}}Hodgkin Disease\\(CUI: C0019829)\end{tabular}} \\
... male with ESRD on \textbf{HD} ...                             & Procedure: Haemodialysis    &                                                                                        &                                                                                     \\
... 3. Asacol \textbf{HD} 800 mg Tablet ... & Part of medication name &                                                                                        &                                                                                     \\
CT scan on \textbf{HD9} showed ...                              & Hospital Day                &                                                                                        &\\
\cline{1-4}
%\textbf{Negation or hypothesis} & \\
%...urinary \textbf{legionella} antigen was also \emph{negative}. & A negated test & \multicolumn{2}{l}{Legionnaires' Disease (CUI: C0023241)} \\
%...\emph{initial concern} for a \textbf{heparin
%induced thrombocytopenia} as his platelet count had... & \multirow{2}{*}{\begin{tabular}[l]{@{}l@{}}A hypothetical disease\end{tabular}} & \multicolumn{2}{l}{\multirow{2}{*}{\begin{tabular}[l]{@{}l@{}}Heparin-induced thrombocytopenia (CUI: C0272285)\end{tabular}}} \\
%\cline{1-4}
\end{tabular}
\end{table*}

\subsection{Weak Supervision}
We propose weak supervision with rules and contextual embeddings to address the above issues of ambiguous mentions. The weak supervision process learns a classifier to decide whether a mention-UMLS pair in the context indicates a correct phenotype of the patient.

The idea is to create rules that allow to select the most reliable subset of mention-UMLS pairs from SemEHR. We propose \textit{mention character length} rule and \textit{``prevalence''} rule, as illustrated in the blue blocks in Fig. \ref{pipeline-weak}. Mention character length rule selects the mentions having over $l$ characters (set as 3 in the pipeline), this helps to remove the abbreviations that are mostly hard to disambiguate, e.g. ``HD'' in Table \ref{fp_example}. The ``prevalence'' rule retains UMLS concepts that represent less than a very small percentage (set as 0.5\%) of the whole mention-UMLS pairs, this is based on the knowledge that rare diseases have low prevalence \cite{scot_gov_2021,textoris_genetic_2014} and thus low mention frequency in the clinical notes. The weak labelling function $f$ is defined as \emph{True} (i.e, mention-UMLS indicates a correct phenotype of the patient) when both rules are satisfied and as \emph{False} when both rules are not satisfied. This omits the data that are only confirmed by one rule, which are not enough reliable to be used for weak supervision. We selected the threshold in both rules to ensure a sufficient amount of reliable, weak data generated.

A \emph{phenotype confirmation model} (to filter the candidate mention-UMLS pairs) can be trained from the weakly labelled data. It is important to ensure that the models do not just learn the rules, but also learn to understand the deeper \textit{context} that can indicate a phenotype mention. We thus propose to use a clinically pre-trained BERT to represent the contexts of the mention. The representation encodes a context window of the mention of size $k$ (including the $k$ tokens before and after the mention, with $k$ set as 5) for classification. Each context window is in a document structure (or a part of the template) of a clinical note. We use the heading of the document structure parsed and named by SemEHR using regular expressions \cite{Wu2018semehr}. To further enrich the contextual information of a mention, we add the document structure before the context window with a separation token [SEP] in between for the representation, similar to the encoding of a question with a paragraph for the Question Answering task in \cite{devlin-etal-2019-bert}. The \emph{contextual representation} is the average of the embeddings (e.g. weights in the second-to-last layer of BERT) of the WordPiece subword tokens of the mention \cite{devlin-etal-2019-bert}, to be fed into a classification model. Instead of fine-tuning the whole BERT model, we use logistic regression as the training model for efficiency, which is similar to adding a feed-forward layer on top of the pre-trained layers (static) in BERT with sigmoid activation for binary classification.

\section{Experiments}
\subsection{Data Processing and Annotation}
Since a large number of rare diseases (especially for the genetic disorders) can lead to an ICU (intensive care unit) admission \cite{textoris_genetic_2014}, we used the discharge summaries (n=59,652) in MIMIC-III (``Medical Information Mart for Intensive Care'') dataset \cite{johnson_mimic-iii_2016}, which contains clinical data from adult patients admitted to the ICU in the Beth Israel Deaconess Medical Center in Boston, Massachusetts between 2001 and 2012. The ICD-9 codes for the admissions also allow us to see how free texts can enrich rare disease coding\footnote{We linked ICD-9 codes to ICD-10 codes using the matching from Ministry of Health, New Zealand \cite{icd9to10newzeland} and linked ICD-9 to UMLS codes based on the ICD-9 ontology in BioPortal \cite{icd9ontology}, as shown in Fig. \ref{pipeline-main}.}. The ontologies and their matchings used are illustrated in Fig. \ref{pipeline-main}.

ORDO contains 14,501 concepts related to rare diseases. We selected the ORDO concepts which have linkage to UMLS and ICD-10 in this study as this supports interoperation among the clinical terminologies, this resulted in a set of 4,064 rare disease concepts.

After processing the discharge summaries with a SemEHR database instance\footnote{\url{https://github.com/CogStack/CogStack-SemEHR}} \cite{Wu2018semehr} with contextual filtering on experiencer and negation, we obtained 127,150 candidate mention-UMLS pairs for the UMLS concepts linked to ORDO. After applying the weak labelling function with the two rules, we finally obtained 15,598 positive data instances and 74,217 negative data instances, and 37,248 non-labelled data (and 87 erroneous data during preprocessing). Each data instance (or mention-UMLS pair) contains the context window of the mention in a document structure (or template section) of a discharge summary and the associated UMLS code for the mention, as shown in Fig. \ref{pipeline-weak}.

For evaluation, we created a gold standard dataset\footnote{The gold standard rare disease mention annotations from the sample of MIMIC-III discharge summaries are available at \url{https://github.com/acadTags/Rare-disease-identification}.} of 1,073 candidate mention-UMLS-ORDO triplets (with each mention in a context window), from a set of randomly sampled 500 discharge summaries from MIMIC-III, where 312 (or 62.5\%) discharge summaries have at least one candidate ``rare disease'' mention. There were in total 95 types of rare disease associated with the mentions. Annotators were asked to label whether a mention-UMLS pair truly indicates a phenotype of the patient. The mention-UMLS pairs were annotated by 3 domain experts, including a research fellow and 2 PhD students in Medical Informatics (MI). Based on the random 200 mention-UMLS pairs annotated by all 3 domain experts, the multi-rater Kappa value was 0.76. ORDO-to-UMLS concept matching was annotated by 2 domain experts (a research fellow and a PhD student in MI) and obtained a Kappa of 0.72. All contradictory and unsure annotations were resolved by a research fellow in biomedical science and MI. We used 400 data instances for model validation and the rest 673 for final testing. Since our pipeline is based on weakly labelled data, none of the gold standard annotations were used for training. Finally, there were 329 (30.7\%) and 146 (13.6\%) correct phenotype mentions in UMLS and ORDO respectively in all 1073 candidate mentions.

\subsection{Implementation Details}
\begin{table*}[ht]
\caption{Evaluation results on Text-to-UMLS linking}
%\footnotesize
\center
\label{umls_filtering_results}
\begin{threeparttable}
\begin{tabular}{llll|lll||lll|lll}
\cline{1-13}
                            & \multicolumn{3}{l}{validation (n=400)}         & \multicolumn{3}{l}{test (n=673)}               & \multicolumn{3}{l}{\begin{tabular}[c]{@{}l@{}}test, \emph{seen} in WS (n=499)\\ i.e. both rules (not) satisfied\end{tabular}} & \multicolumn{3}{l}{\begin{tabular}[c]{@{}l@{}}test, \emph{unseen} in WS (n=174)\\i.e. only one rule satisfied \end{tabular}} \\
Text-to-UMLS linking      & P             & R              & $F_1$         & P             & R              & $F_1$         & P                  & R                   & $F_1$             & P                  & R                   & $F_1$             \\
\cline{1-13}
GHNL API \cite{Bodnari2020} & 78.9          & 81.7           & 80.3          & 75.3          & 78.1           & 76.6          & 54.3               & 62.5                & 58.1 & \textbf{94.1}      & 89.7                & \textbf{91.9}                   \\ \cline{1-1} \cline{1-13}
SemEHR* \cite{Wu2018semehr} & 35.5          & \textbf{100.0} & 52.4          & 27.8          & \textbf{100.0} & 43.5          & 16.0               & \textbf{100.0}      & 27.6 & 61.5               & \textbf{100.0}      & 76.1                            \\
+ rules                     & 80.9          & 89.4           & 84.9          & 68.6          & 94.7           & 79.6          & \textbf{83.3}      & 87.5                & \textbf{85.4} & 61.5               & \textbf{100.0}      & 76.1                   \\
+ WS (rules+BlueBERT, ours)                 & \textbf{90.1} & 89.4           & \textbf{89.8} & \textbf{80.4} & 92.0           & \textbf{85.8} & \textbf{83.3}      & 87.5                & \textbf{85.4} & 78.5               & 95.3       & 86.1                   \\
\cline{1-13}
\end{tabular}
\begin{tablenotes}
\item$^{*}$ SemEHR has a perfect reference recall, because all candidate mention-UMLS pairs were originally created using the tool.
\end{tablenotes}
\end{threeparttable}
\end{table*}

We used the open-source tool, bert-as-service\footnote{\url{https://github.com/hanxiao/bert-as-service}} \cite{xiao2019bertservice}, built on Google AI's BERT implementation with Python Tensorflow\footnote{\url{https://github.com/google-research/bert}} \cite{devlin-etal-2019-bert}, to encode each mention in a raw contextual window as a vector representation. Specifically, we used the BlueBERT \cite{peng2019transfer} model pre-trained on MIMIC-III clinical notes and PubMed abstracts. We then trained a logistic regression model with the representations, with default configuration using the Python scikit-learn package \cite{scikit-learn}, on the weakly labelled mention-UMLS pairs.

We evaluated the approach using precision, recall, and $F_1$ scores. We compared the proposed approach (SemEHR+WS) with SemEHR with the two rules only using an OR operation (SemEHR+rules), where no weak supervision was performed. To note that SemEHR had a reference recall of 100\% as all candidate ``rare disease'' mentions were identified by SemEHR, which was a source of the annotations.

We also compared the results to a third-party tool, Google Healthcare Natural Language API (GHNL API)\footnote{\url{https://cloud.google.com/healthcare/docs/concepts/nlp}.}, released on Nov 2020 \cite{Bodnari2020}. Similar to SemEHR, GHNL API identifies clinical entities from texts and links them to UMLS and other ontologies. We assume that the mention-UMLS pair is predicted as \emph{True} if the same UMLS concept is detected by the GHNL API from (a part of) the mention after contextual filtering (with ``certainly assessment'' and ``subject'' values).%\footnote{We selected the entities with certainty assessment no less than ``SOMEWHAT LIKELY'' and subject as ``PATIENT'', see \url{https://cloud.google.com/healthcare/docs/concepts/nlp\#supported_functional_feature_categories}.}).

\subsection{Results}
Table \ref{umls_filtering_results} shows Text-to-UMLS linking results. It can be observed that with weak supervision (WS), the precision and $F_1$ has significantly improved by over 50\% and 40\% absolute value respectively compared to SemEHR. Adding the two customised rules can already improve the testing performance greatly by around 35\% testing $F_1$ to SemEHR (as shown in SemEHR+rules), which validates the efficiency of the proposed rules with the entity linking tool to create reliable weak annotations. WS further outperformed the rule-enhanced approach absolutely by around 10\% precision (and 5\% $F_1$), showing that the BERT representation helps the model to filter the false positives by representing the contexts of a mention. The recall dropped slightly by introducing the two rules with SemEHR, this suggests that the rules, while being effective for WS, can still filter out true positive mentions. Also with WS, the overall approach significantly outperforms the clinical entity extraction tool, GHNL API, by over 9\% $F_1$. These results together suggest that the proposed WS approach can successfully adapt and improve a gazetteer-based tool to a specific domain (e.g. rare diseases).

To analyse the impact of weak supervision on the testing data, we split the testing data into those ``\emph{seen}'' or ``\emph{unseen}'' during weak supervision, based on the rules and weak labelling function. ``Seen'' data mean that the mention-UMLS pairs were weakly labelled with the function $f$, i.e. with both rules satisfied or both not satisfied; ``unseen'' data mean that only one of the rules was satisfied, thus the data were not labelled. WS improved the performance of SemEHR in both settings. While the weakly ``seen'' data were dramatically boosted by rules (by nearly 50\% $F_1$), the ``unseen'' data were significantly improved (by 10\% $F_1$) through the model generalised with contextual representations. We also see that the GHNL API achieved even better precision and $F_1$ for the weakly ``unseen'' data (and worse for the ``seen'' data).

\begin{table}[ht]
\caption{Results on rare disease identification (Text-to-ORDO)}
\center
\label{overall_rd_id_results}
\begin{tabular}{llll}
\cline{1-4}
                 & \multicolumn{3}{l}{evaluation set (n=1,073)} \\
Text-to-ORDO linking               & P           & R           & $F_1$        \\
\cline{1-4}
GHNL API \cite{Bodnari2020}         & 45.8        & 55.5        & 50.2        \\
\cline{1-4}
SemEHR \cite{Wu2018semehr}           & 15.7        & \textbf{93.8}        & 26.9        \\
+ rules          & 50.9        & 81.5        & 62.6        \\
+ WS (rules+BlueBERT, ours) & \textbf{63.7}        & 78.1        & \textbf{70.2}        \\
\cline{1-4}
\end{tabular}
\end{table}

Combining the Text-to-UMLS and UMLS-to-ORDO\footnote{The UMLS-to-ORDO matching accuracy for the 95 concept pairs was improved by 1.0\% (from 87.4\% to 88.4\%) with semantic type filtering.} modules, the overall rare disease identification results are presented in Table \ref{overall_rd_id_results}. For the whole 1073 evaluation data, our approach achieved significantly better precision and $F_1$ compared to other methods, with 7.6\% $F_1$ above SemEHR with rules only, and 20\% $F_1$ better than the GHNL API.

While rules are effective for WS, they may also introduce some errors. Our brief analysis of the overall approach shows that over half (52.2\%) of the 67 errors from the Text-to-UMLS side were due to the bias introduced from the weak rules. Mentions with negated or hypothetical contexts (represented 17.9\% of the errors) were also challenging for the algorithm and the annotators. Both issues may be addressed by combining WS with human-in-the-loop machine learning \cite{monarch2021} with adaptive rules to improve the performance.

\begin{figure}
  \centering
  \includegraphics[width=0.48\textwidth]{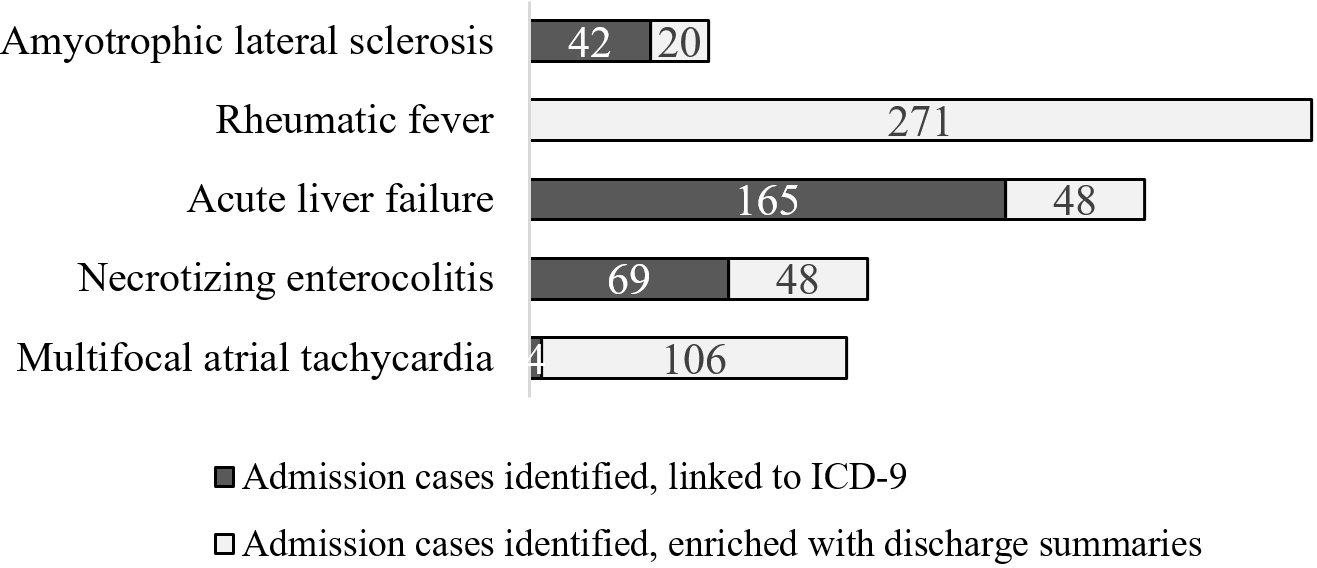}
  \caption{Number of rare disease admission cases identified from \emph{discharge summaries} (n=59,652) in MIMIC-III. The 5 rare diseases of perfect admission-level evaluation $F_1$ and most number of cases in the evaluation data are presented. The cases are split into those linked to the admission ICD-9 codes and those further enriched with free texts.}\label{text_vs_icd_fig}
\end{figure}

After processing all discharge summaries with the overall pipeline, the overall system finally identified 10,488 (17.6\% of all 59,652) ICU admissions in MIMIC-III associated with at least one rare disease. There were in total 466 types of rare disease identified. Through linking admission ICD-9 codes to the ORDO concepts for each admission (see dotted lines in Fig. \ref{pipeline-main}), we observed that 92.5\% (431/466) of the rare diseases were potentially under-coded for at least one admission. Fig. \ref{text_vs_icd_fig} shows the admission cases identified for 5 selected rare disease cases (all with perfect admission-level $F_1$ on the evaluation set). Some rare diseases are likely heavily under-coded, e.g. none (0\%) of the 271 Rheumatic Fever cases and only 4 (3.6\%) of the 110 (=4+106) Multifocal atrial tachycardia cases were coded. This shows the necessity of using free texts to enrich rare disease cohort selection.

\section{Conclusion}
We proposed a practical and effective ontology-based and weak supervision approach to leverage a named entity linking tool, SemEHR, for rare disease identification from clinical notes. The results show that weak supervision with customised rules and contextual representations can greatly improve the performance, without the annotation from domain experts. The rare disease cases identified from discharge summaries can greatly enrich cases identified using the admission ICD codes. The applied tool, rules, and ontologies can be adapted and extended to extract information from other documents. The study was performed using discharge summaries from ICU in a US hospital. A future study is to externally validate the approach to reports in other clinical domains and institutions (e.g. radiology reports in the UK).
%\addtolength{\textheight}{-12cm}   % This command serves to balance the column lengths
                                  % on the last page of the document manually. It shortens
                                  % the textheight of the last page by a suitable amount.
                                  % This command does not take effect until the next page
                                  % so it should come on the page before the last. Make
                                  % sure that you do not shorten the textheight too much.

\section*{Acknowledgement}

The authors would like to thank the comments from Dr William Whiteley and other members in the Clinical Natural Language Processing Research Group, University of Edinburgh. This work has also made use of the resources provided by the Edinburgh Compute and Data Facility (ECDF).

\bibliography{rare_disease_coding_res}

% Generated by IEEEtran.bst, version: 1.14 (2015/08/26)
\begin{thebibliography}{10}
\providecommand{\url}[1]{#1}
\csname url@samestyle\endcsname
\providecommand{\newblock}{\relax}
\providecommand{\bibinfo}[2]{#2}
\providecommand{\BIBentrySTDinterwordspacing}{\spaceskip=0pt\relax}
\providecommand{\BIBentryALTinterwordstretchfactor}{4}
\providecommand{\BIBentryALTinterwordspacing}{\spaceskip=\fontdimen2\font plus
\BIBentryALTinterwordstretchfactor\fontdimen3\font minus
  \fontdimen4\font\relax}
\providecommand{\BIBforeignlanguage}[2]{{%
\expandafter\ifx\csname l@#1\endcsname\relax
\typeout{** WARNING: IEEEtran.bst: No hyphenation pattern has been}%
\typeout{** loaded for the language `#1'. Using the pattern for}%
\typeout{** the default language instead.}%
\else
\language=\csname l@#1\endcsname
\fi
#2}}
\providecommand{\BIBdecl}{\relax}
\BIBdecl

\bibitem{scot_gov_2021}
{Scottish Government}, \emph{Illnesses and long-term conditions}, 2021,
  \url{https://www.gov.scot/policies/illnesses-and-long-term-conditions/rare-diseases/}
  (accessed Mar. 22, 2021).

\bibitem{Bearryman2016}
E.~Bearryman, \emph{Does your rare disease have a code?}, 2016,
  \url{https://www.eurordis.org/news/does-your-rare-disease-have-code}
  (accessed July 29, 2021).

\bibitem{rios-kavuluru-2018-shot}
A.~Rios and R.~Kavuluru, ``Few-shot and zero-shot multi-label learning for
  structured label spaces,'' in \emph{Proceedings of the 2018 Conference on
  Empirical Methods in Natural Language Processing}.\hskip 1em plus 0.5em minus
  0.4em\relax Brussels, Belgium: Association for Computational Linguistics,
  Oct.-Nov. 2018, pp. 3132--3142.

\bibitem{dong2021}
H.~Dong, V.~Suárez-Paniagua, W.~Whiteley, and H.~Wu, ``Explainable automated
  coding of clinical notes using hierarchical label-wise attention networks and
  label embedding initialisation,'' \emph{Journal of Biomedical Informatics},
  p. 103728, 2021.

\bibitem{searle2020}
T.~Searle, Z.~Ibrahim, and R.~Dobson, ``Experimental evaluation and development
  of a silver-standard for the {MIMIC}-{III} clinical coding dataset,'' in
  \emph{Proceedings of the 19th SIGBioMed Workshop on Biomedical Language
  Processing}.\hskip 1em plus 0.5em minus 0.4em\relax Online: Association for
  Computational Linguistics, Jul. 2020, pp. 76--85.

\bibitem{kahn2017ontology}
C.~E. Kahn~Jr, ``An ontology-based approach to estimate the frequency of rare
  diseases in narrative-text radiology reports.'' \emph{Studies in Health
  Technology and Informatics}, vol. 245, pp. 896--900, 2017, mEDINFO 2017:
  Precision Healthcare through Informatics.

\bibitem{weinreich2008orphanet}
S.~S. Weinreich, R.~Mangon, J.~Sikkens, M.~Teeuw, and M.~Cornel, ``Orphanet: a
  european database for rare diseases,'' \emph{Nederlands tijdschrift voor
  geneeskunde}, vol. 152, no.~9, pp. 518--519, 2008.

\bibitem{wang_clinical_2019}
Y.~Wang, S.~Sohn, S.~Liu, F.~Shen, L.~Wang, E.~J. Atkinson, S.~Amin, and
  H.~Liu, ``A clinical text classification paradigm using weak supervision and
  deep representation,'' \emph{BMC Medical Informatics and Decision Making},
  vol.~19, no.~1, p.~1, Jan. 2019.

\bibitem{ratner2019}
A.~Ratner, P.~Varma, B.~Hancock, C.~Ré, and other members~of Hazy~Lab,
  \emph{Weak Supervision: A New Programming Paradigm for Machine Learning},
  2019, \url{http://ai.stanford.edu/blog/weak-supervision/} (accessed Mar. 13,
  2021).

\bibitem{devlin-etal-2019-bert}
J.~Devlin, M.-W. Chang, K.~Lee, and K.~Toutanova, ``{BERT}: Pre-training of
  deep bidirectional transformers for language understanding,'' in
  \emph{Proceedings of the 2019 Conference of the North {A}merican Chapter of
  the Association for Computational Linguistics: Human Language Technologies,
  Volume 1 (Long and Short Papers)}.\hskip 1em plus 0.5em minus 0.4em\relax
  Minneapolis, Minnesota: Association for Computational Linguistics, Jun. 2019,
  pp. 4171--4186.

\bibitem{cusick2021}
M.~Cusick, P.~Adekkanattu, T.~R. Campion, E.~T. Sholle, A.~Myers, S.~Banerjee,
  G.~Alexopoulos, Y.~Wang, and J.~Pathak, ``Using weak supervision and deep
  learning to classify clinical notes for identification of current suicidal
  ideation,'' \emph{Journal of Psychiatric Research}, vol. 136, pp. 95--102,
  2021.

\bibitem{shen2021}
Z.~Shen, Y.~Yi, A.~Bompelli, F.~Yu, Y.~Wang, and R.~Zhang, ``Extracting
  lifestyle factors for alzheimer's disease from clinical notes using deep
  learning with weak supervision,'' \emph{arXiv preprint arXiv:2101.09244},
  2021.

\bibitem{Wu2018semehr}
H.~Wu, G.~Toti, K.~I. Morley, Z.~M. Ibrahim, A.~Folarin, R.~Jackson,
  I.~Kartoglu, A.~Agrawal, C.~Stringer, D.~Gale, G.~Gorrell, A.~Roberts,
  M.~Broadbent, R.~Stewart, and R.~J. Dobson, ``{Sem{EHR}: A general-purpose
  semantic search system to surface semantic data from clinical notes for
  tailored care, trial recruitment, and clinical research},'' \emph{Journal of
  the American Medical Informatics Association}, vol.~25, no.~5, pp. 530--537,
  01 2018.

\bibitem{peng2019transfer}
Y.~Peng, S.~Yan, and Z.~Lu, ``Transfer learning in biomedical natural language
  processing: An evaluation of bert and elmo on ten benchmarking datasets,'' in
  \emph{Proceedings of the 2019 Workshop on Biomedical Natural Language
  Processing (BioNLP 2019)}, 2019, pp. 58--65.

\bibitem{johnson_mimic-iii_2016}
A.~E.~W. Johnson, T.~J. Pollard, L.~Shen, L.-w.~H. Lehman, M.~Feng,
  M.~Ghassemi, B.~Moody, P.~Szolovits, L.~A. Celi, and R.~G. Mark,
  ``{MIMIC}-{III}, a freely accessible critical care database,''
  \emph{Scientific Data}, vol.~3, no.~1, pp. 1--9, May 2016.

\bibitem{Bodnari2020}
A.~Bodnari, \emph{Healthcare gets more productive with new industry-specific AI
  tools}, 2020,
  \url{https://cloud.google.com/blog/topics/healthcare-life-sciences/now-in-preview-healthcare-natural-language-api-and-automl-entity-extraction-for-healthcare}
  (accessed Mar. 15, 2021).

\bibitem{gorrell2018}
G.~Gorrell, X.~Song, and A.~Roberts, ``Bio-yodie: A named entity linking system
  for biomedical text,'' \emph{arXiv preprint arXiv:1811.04860}, 2018.

\bibitem{textoris_genetic_2014}
\BIBentryALTinterwordspacing
J.~Textoris and M.~Leone, ``\BIBforeignlanguage{en}{Genetic {Aspects} of
  {Uncommon} {Diseases}},'' in \emph{\BIBforeignlanguage{en}{Uncommon
  {Diseases} in the {ICU}}}, M.~Leone, C.~Martin, and J.-L. Vincent, Eds.\hskip
  1em plus 0.5em minus 0.4em\relax Cham: Springer International Publishing,
  2014, pp. 3--11. [Online]. Available:
  \url{https://doi.org/10.1007/978-3-319-04576-4_1}
\BIBentrySTDinterwordspacing

\bibitem{icd9to10newzeland}
{Ministry of Health NZ}, \emph{Mapping between ICD-10 and ICD-9}, 2000,
  \url{https://www.health.govt.nz/nz-health-statistics/data-references/mapping-tools/mapping-between-icd-10-and-icd-9}
  (accessed Apr. 30, 2021).

\bibitem{icd9ontology}
{NCBO BioPortal}, \emph{International Classification of Diseases, Version 9 -
  Clinical Modification}, 2021,
  \url{https://bioportal.bioontology.org/ontologies/ICD9CM} (accessed Apr. 30,
  2021).

\bibitem{xiao2019bertservice}
H.~Xiao, \emph{Serving Google BERT in Production using Tensorflow and ZeroMQ},
  2019,
  \url{https://hanxiao.io/2019/01/02/Serving-Google-BERT-in-Production-using-Tensorflow-and-ZeroMQ/}
  (accessed Apr. 25, 2021).

\bibitem{scikit-learn}
F.~Pedregosa, G.~Varoquaux, A.~Gramfort, V.~Michel, B.~Thirion, O.~Grisel,
  M.~Blondel, P.~Prettenhofer, R.~Weiss, V.~Dubourg, J.~Vanderplas, A.~Passos,
  D.~Cournapeau, M.~Brucher, M.~Perrot, and E.~Duchesnay, ``Scikit-learn:
  Machine learning in {P}ython,'' \emph{Journal of Machine Learning Research},
  vol.~12, pp. 2825--2830, 2011.

\bibitem{monarch2021}
R.~M. Monarch, \emph{Human-in-the-Loop Machine Learning: Active learning and
  annotation for human-centered AI}.\hskip 1em plus 0.5em minus 0.4em\relax
  Shelter Island, NY: Manning Publications Company, 2021, version 11, MEAP
  Edition (Manning Early Access Program).

\end{thebibliography}
\bibliographystyle{IEEEtran}

\end{document}